\documentclass[11pt]{article}

\usepackage[a4paper, total={6.5in, 9.5in}]{geometry}

\usepackage{times}
\usepackage{color}
\usepackage{amsfonts}
\usepackage{graphicx}
\usepackage{subcaption}
\usepackage{amsmath,amsfonts,amssymb,bm}
\usepackage{booktabs} 
\usepackage{array}
\usepackage{boxedminipage}
\usepackage{paralist}
\usepackage{url}
\usepackage{authblk}
\usepackage{pbox}

\usepackage{diagbox}
\usepackage{enumitem}
\usepackage{commath}
\usepackage{paralist}
\definecolor{darkspringgreen}{rgb}{0.05, 0.5, 0.06}
\usepackage{arydshln}
\definecolor{Gray}{gray}{0.8}

\newcolumntype{L}[1]{>{\centering\let\newline\\\arraybackslash\hspace{0pt}}m{#1}}
\newcolumntype{C}[1]{>{\centering\let\newline\\\arraybackslash\hspace{0pt}}m{#1}}
\newcolumntype{R}[1]{>{\centering\let\newline\\\arraybackslash\hspace{0pt}}m{#1}}

\newcommand{\argmax}{\arg\!\max}

\title{Monolingual sentence matching for text simplification}

\begin{document}
\author{Yonghui Huang*, Yunhui Li*, Yi Luan**  \\*Harbin University\\ **University of Washington }
\title{Monolingual sentence matching for text simplification}
\date{}
\maketitle


%

\begin{abstract}
This work improves monolingual sentence alignment for text simplification, specifically for text in standard and simple Wikipedia. We introduce a convolutional neural network structure to model similarity between two sentences.  Due to the limitation of available parallel corpora, the model is trained in a semi-supervised way, by using the output of a knowledge-based high performance aligning system. We apply the resulting similarity score to rescore the knowledge-based output, and adapt the model by a small hand-aligned dataset. Experiments show that both rescoring and adaptation improve the performance of knowledge-based method. 
\end{abstract}
\section{Introduction}
Text simplification is an operation in natural language processing to modify an existing corpus of human-readable text so that the grammar and structure of the sentence is greatly simplified, while the underlying meaning and information remains the same. Due to the fact that natural human languages ordinarily contain complex compound constructions~\cite{luan2017scienceie}, text simplification can make texts easier for human readers as well as automatic text processing. Although there are many previous works regarding text simplification e.g.(\cite{callison}, \cite{fung}), which benifit from data-driven machine translation, paraphrasing or grounded language acquisition techniques, works are still limited because available monolingual parallel corpora are limited or automatically generated are noisy. 

Wikipedia is potentially a good resource for text simplification, since it includes standard articles and their corresponding simple articles in English.  A challenge with automatic alignment is that standard and simple articles can be written independently so they are not strictly parallel, and have very different presentation ordering. A few studies use editor comments attached to Wikipedia edit logs to extract pairs of simple and difficult words \cite{yatskar}. Works such as \cite{zhu}, use text-based similarity techniques to extract pairs of simple and standard sentences. \cite{sim} use a greedy search over the simple and standard Wikipedia documents to align sentence pairs. By taking advantage of a word-level semantic similarity measure built on top of Wiktionary and WordNet, \cite{sim} do not make assumption about the relative order of sentences in standard versus simple Wikipedia articles and obtain the highest performance. The resulting datasets of manually and automatically aligned sentence pairs are made available. This paper uses the automatic aligned sentence pairs of \cite{sim} to do semi-supervise training and manually aligned data to do adaptation and performance evaluation. 

Based on this setting of using Wikipedia on text simplification, the goal of this project is to find sentence level alignment. The problem is solved by first matching a sentence to its correponding simple version. In order to capture rich contextual and semantic information in a sentence, we use a Convolutional Neural Network (CNN) structure to learn low-dimensional, semantic vector representation for sentence. CNN is proven successful on image\cite{image} and speech \cite{speech}. In previous studies, convolutional notion of similarity is efficient in capturing contenxtual information and has been vastly applied to natural language sentence matching. In \cite{clsm}, a Convolutional Latent Semantic Model (CLSM) is proposed to model information retrieval, which use one global max-pooling layer and a deep strucure to extract fixed length sentence-level embedding. In \cite{match} a deeper and more interactive convolutional structure is proposed to matching sentences, by applying multiple maxpooling and convolutional layers~\cite{luan2016multiplicative}.  The objective function of these models is to maximaze the similarity score between two sentence embeddings againest several negative samples, which is flexible in fitting in different tasks~\cite{luan2016lstm}. Tasks such as paraphrase identification, information retrieval have shown great improvement when applying this deep convolutional structures. Meanwhile, \cite{word2veccnn} trained a simple CNN with one layer of convolution on top of word embeddings obtained from an unsupervised neural language model \cite{word2vec,luan2018multi} and obtained excellent performance in sentence classification when fine tuning the pre-trained word vectors. Therefore, in this paper, we introduce pre-trained word embeddings and fine-tuning into CLSM and compare its performance with traditional letter-trigram CLSM on text simplification task. Our approach gives comparable performance as knowledge-based methods in \cite{sim}. Since the model is pre-trained, the approach is much faster when testing.

CLSM starts with each word within a temporal context window in a word sequence to directly capture contextual features at n-gram levels. Next, salient word n-gram features are extracted and then aggregated to form a sentence-level vector. Sentence level similarity can be calculated by cosine distance of vector representation of parallel sentences. CLSM fits well to our goal due to two reasons. First of all, sentence level alignment data can be used to train CLSM, then we can use CLSM to find similarity between new sentence pairs. Secondly, on max-pooling layer, salient feature are extracted. When tracing back to the neurons at max-pooling layer, n-grams that have high activation values for both sentences would have semantic matching, which could provide information of phrase-level matching, which is our next-step direction~\cite{luan2015efficient,luan2014relating,luan2017multi}. 

The reminder of the report is organized as follows. In section 2, CLSM and its application to text simplification is briefly introduced. In section 3, the results of experiments is described. In section 4, experimental results are analyzed, insights for extracting phrase-level matches are described. Finally, Section 5 draws overall conclusions and describes possible future work.

\section{The CLSM Architecture}
\subsection{Letter-trigram based word n-gram representation}
In order to reduce vocabulary size and reduce out-of-vocabulary problem, we use word hashing technique, which represents a word by a letter-trigram. For example, given a word 'boy', the word is represented by '\#-b-o', 'b-o-y', 'o-y-\#'. The \textit{t} th word is represented as a count vector of letter trigrams $f_{t}$. In Figure \ref{1}, the letter trigram matrix $W_f$ denotes the letter-trigram transformation.  Given the letter-trigram representation, each word trigram is represented by
\begin{equation}
l_t = [f_{t-1}^{T}, ..., f_{t}^{T}, ..., f_{t+1}^{T}], \quad t=1,...,T
\end{equation}

One problem of this method is collision, i.e., two different words could have the same letter n-gram vector representation. According to \cite{DSSM}, 
500K-word vocabulary can be represented by a 30,621 dimensional vector using letter trigrams, a reduction of 16-fold in
dimensionality with a negligible collision rate of 0.0044\%.

 While the number of English words can be unlimited, the number of letter n-grams in English (or other similar languages) is often limited. Moreover, word hashing is able to map the morphological variations of the same word to the points that are close to each other in the letter n-gram space. More importantly, while a word unseen in the training set always cause difficulties in word-based representations, it is not the case where the letter n-gram
based representation is used. Thus, letter n-gram based word hashing is robust to the out-of-vocabulary problem.

\subsection{Word embedding initialization and fine-tuning}

Initializing word vectors with word embeddings obtained from an unsupervised neural language model is a popular method to improve performance in the absence of a large supervised training set (\cite{socher}, \cite{collobert}). One advantage of word embedding initialization compared to letter-trigram or bag of words is that it can significantly reduce feature dimension which could prevent overfitting and reduce computational cost.  Word embedding initialization from unsupervised language model could introduce prior knowledge of the dataset, yet introduce mismatch for the specific training task at the same time. One way of solving the mismatching problem is to task specificaly fine-tune the word embeddings, which shows further gain in many previous researches \cite{word2veccnn}.

\begin{figure}[tb]
\centering
\includegraphics[width=14cm]{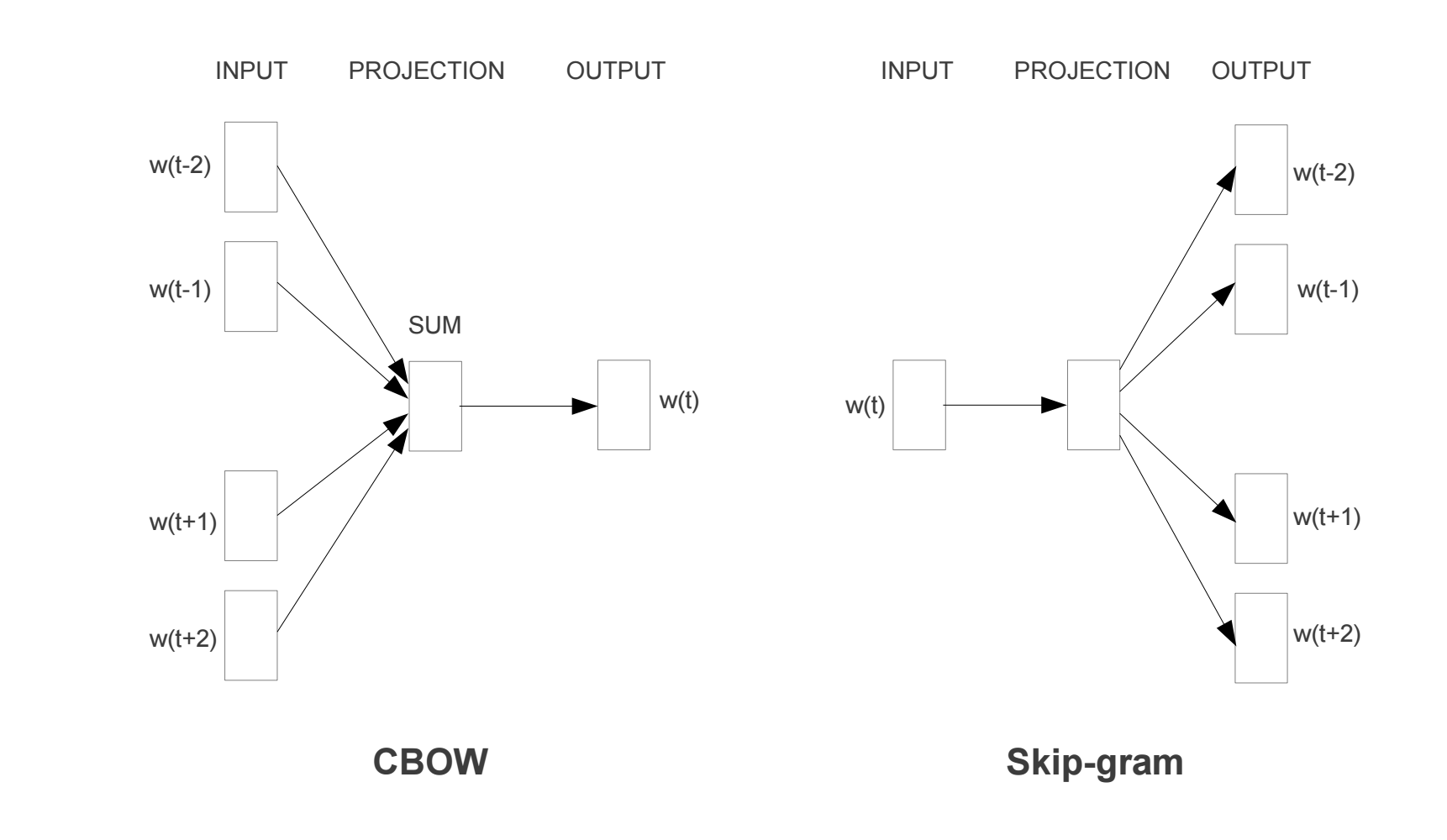}
\caption{{\it Two word2vec architectures, figure from \cite{word2vec}}}
\label{word2vec}
\end{figure}

Among different word embedding approaches, \textit{word2vec}\cite{word2vec} has been widely applied to different NLP tasks, and proved great performance in word similarity task. The code for \textit{word2vec} is publicly available and can be trained efficiently from large corpus of natural language. \textit{word2vec} has two structures: Continuous Bag-of-Words Model (CBOW) and Continuous Skip-gram Model (Skip-gram), as Figure \ref{word2vec}. CBOW predicts the current word based on the context, and the Skip-gram predicts surrounding words given the current word. We test both structure and use Skip-gram since it gives best performance. In order to compare the performance of word embedding and letter-trigram, the context window remains the same across different experiments. 

\subsection{Convolutional layer, max pooling layer and latent semantic layer}

The convolutional operation\cite{luan2018uwnlp} can be viewed as sliding window based feature extraction. It is designed to capture the word n-gram contextual features. Consider the \textit{t}-th word n-gram, the convolution matrix projects its letter-trigram representation vector $l_t$ to a contextual feature vector $h_t$, $h_t$ is computed by
\begin{equation}
h_t = tanh(W_c \cdot l_t), \quad t=1,...,T
\end{equation}
where $W_c$ is the feature transformation matrix, as known as the convolution matrix, that are shared among all word n-grams. $tanh$ is used as the activation function of the neurons:
\begin{equation}
tanh(x) = \frac{1-e^{-2x}}{1+e^{-2x}}
\end{equation}

\begin{figure}[tb]
\centering
\includegraphics[width=8cm]{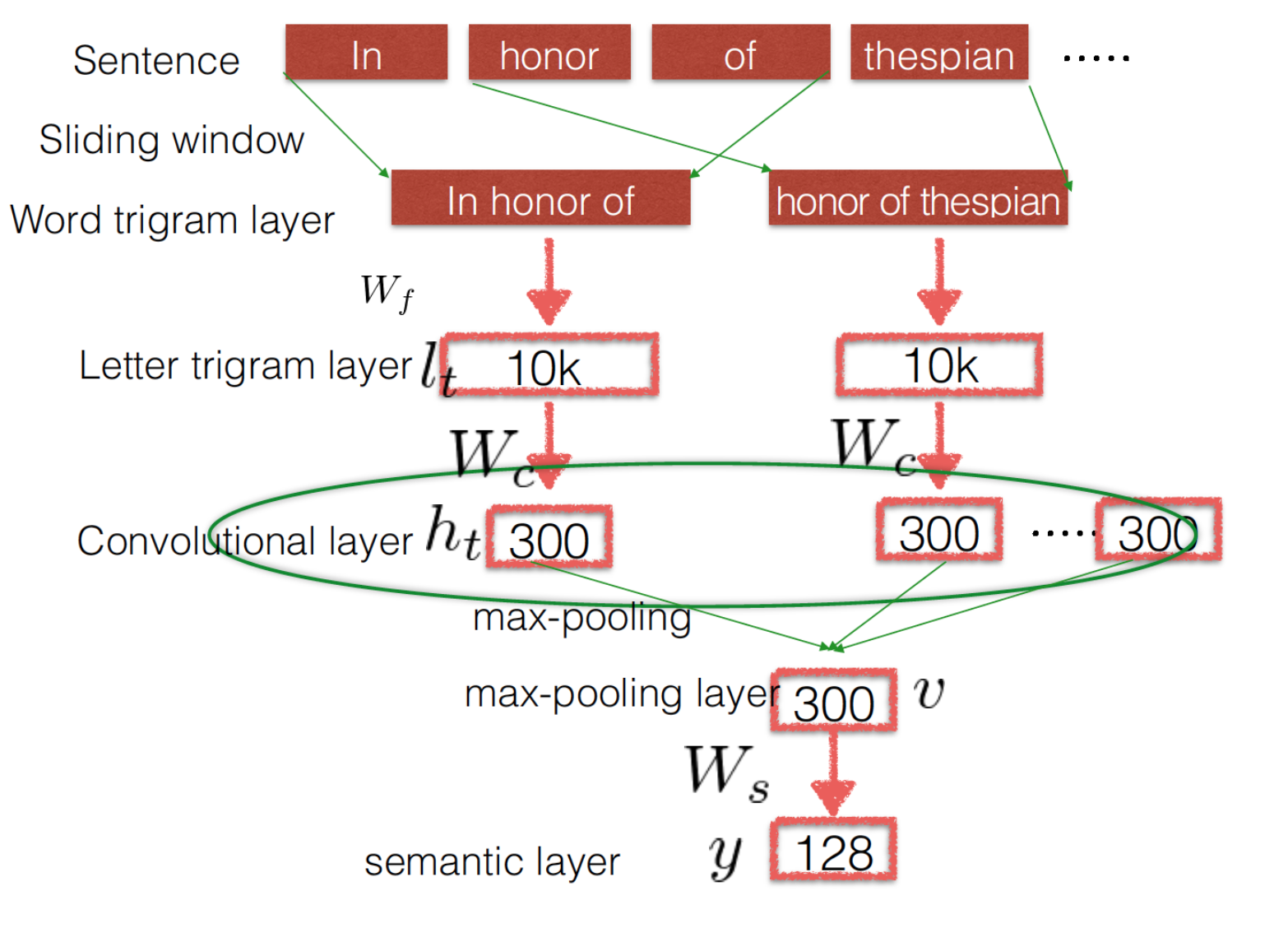}
\caption{{\it The CLSM structure}}
\label{1}
\end{figure}

In order to aggregate the salient word into sentence-level feature vector, CLSM use max pooling to force the network to retain only the most useful local features produced by convolutional layers. As Figure \ref{1}, the max-pooling layer is 
\begin{equation}
v(i) = max_{t=1,...,T}\{h_t(i)\}, \quad i=1,...,K
\end{equation}
where $v(i)$ is the \textit{i}-th element of the max pooling layer $v$, $h_t(i)$ is the \textit{i}-the element of the \textit{t}-th local feature vector $h_t$. $K$ is the dimensionality of the max pooling layer, which is the same as the dimensionality of the local contextual feature vectors \{$h_t$\}.

After the sentence-level feature is produced by the max-pooling operation, one more non-linear transformation layer is applied to extract the high-level semantic representation, denoted by $y$. As shown in Figure \ref{1}, we have
\begin{equation}
y=tanh(W_s\cdot v)
\end{equation}

where $v$ is the global feature vector after max pooling, $W_s$ is the semantic projection matrix, and y is the vector representation of the sentence in the latent semantic space, with a dimensionality of $L$.

\subsection{Using CLSM for sentence similarity modeling}
We use cosine similarity to measure the similarity between two sentences. The score between a standard and a simple sentence is defined as:
\begin{equation}
\label{similarity}
R(sim, std)=cos(y_{sim},y_{std}) = \frac{y_{sim}^T y_{std}}{||y_{sim}||||y_{std}||}
\end{equation}

where $y_{sim}$ and $y_{std}$ are the semantic vector of the simple sentence and the standard sentence, respectively.

The data for training CLSM is the parallel simple and standard sentences from Wikipedia. CLSM is trained in a way that the score of matched sentences are maximized. In order to align sentences in simple document with sentences in standard document, for each simple sentence, 4 randomly selected standard sentences (matched sentence excluded) is collected together as its matched standard sentence.  Therefore, the probability of the simple sentence paired with its matched standard sentence is given through softmax:
\begin{equation}
P(std^{+}|sim)=\frac{\exp{(R(sim, std^{+}))}}{\sum_{std^{'}\in\bm{std}}\exp(R(sim, std^{'}))} 
\end{equation}

Likewise, the probability of the standard sentence paired with its matched simple sentence is given through softmax:

\begin{equation}
P(sim^{+}|std)=\frac{\exp{(R(std, sim^{+}))}}{\sum_{sim^{'}\in\bm{sim}}\exp(R(std, sim^{'}))} 
\end{equation}

In training, the model parameters are learned to maximize the likelihood of the best match standard sentences given the target simple sentences and the best simple sentences given the target simple sentences. That is, we minimize the following loss function
\begin{equation}
L(\Lambda) = -\log \prod_{(sim, std^{+})}P(std^{+}|sim)  \prod_{(std, sim^{+})}P(sim^{+}|std)
\end{equation}

\section{Baseline methods}
As a previous study, this work use the same dataset and evualuation metrics as \cite{sim}, which contains a sentence-level similarity score that builds
on a new word-level semantic similarity, described
below, together with a greedy search over the article.
\subsection{Word-level similarity}
\label{baseline}
Word-level similarity functions return a similarity
score $\sigma(w_1, w_2)$ between words $w_1$ and $w_2$. \cite{sim} introduces
a novel similarity metric: structural semantic similarity. The paper compare Wordnet structural semantic similarity scores \cite{wordnet} with Wiktionary structural semantic similarity scores. We therefore use these two methods as our baseline. For details of extracting Wordnet and Wiktionary similarity scores, please refer to \cite{sim} and \cite{wordnet}.
\subsection{Structural semantic similarity}
\cite{sim} extend the word-level similarity metric to account for both semantic
similarity between words, as well as the
dependency structure between the words in a sentence.
A triplet is created for each word using Stanford’s
dependency parser \cite{parser}.
Each triplet $t_w = (w, h, r)$ consists of the given
word $w$, its head word $h$ (governor), and the dependency
relationship (e.g., modifier, subject, etc)
between $w$ and $h$. The similarity between words
$w_1$ and $w_2$ combines the similarity between these
three features in order to boost the similarity score
of words whose head words are similar and appear
in the same dependency structure: $\sigma_{ss_{wk}} (w_1, w_2) =
\sigma_{w_k}(w_1, w_2) + \sigma_{w_k}(h_1, h_2)\sigma_r(r_1, r_2)$ where $\sigma{w_k}$ is
the WikNet similarity and $\sigma_{r}(r_1, r_2)$ represents dependency
similarity between relations $r_1$ and $r_2$
such that $\sigma_r = 0.5$ if both relations fall into the same
category, otherwise $\sigma_r = 0$.
\subsection{Greedy sequence-level alignment}
\label{greedy}
To avoid aligning multiple sentences to the same
content, one-to-one matches is required between
sentences in simple and standard Wikipedia articles
using a greedy algorithm. Similarities
between all sentences $S_{std}^{(j)}$ in the simple article
and $S_{sim}^{(i)}$ in standard article using a sentence level
similarity score is first computed. Then, iteratively
selects the most similar sentence pair $S_{std}^{*}, S_{sim}^{*} =\argmax s(S_{std}^{(j)} , S_{sim}^{(i)}) $ and removes all other pairs associated
with the respective sentences, repeating until
all sentences in the shorter document are aligned.

\section{Experiments}
\subsection{Data sets and experimental setup}
The data for experiments is from \cite{sim}. Both manually and automatically
aligned sentence pairs are available.
The manually annotated dataset is used as a test set for
evaluating alignment methods as well as tuning parameters
for generating automatically aligned pairs
across standard and simple Wikipedia.
46 article pairs are randomly selected from Wikipedia (downloaded in June
2012) that started with the character 'a'. The annotators were given three choices: Good (complete match in semantics), Good partial (contains additional clause or phrase), Partial (discuss unrelated concepts, but share a short related phrase) and Bad (completely unrelated concepts). 
The whole dataset include 67853 sentence pairs (277 good, 281 good partial, 117 partial and 67178 bad). The
kappa value for interannotator agreement is 0.68
(13\% of articles were dual annotated). Most disagreements
between annotators are confusions between
‘partial’ and ‘good partial’ matches. An example of good, good partial, partial pairs is listd in Table \ref{good}.

\cite{sim} release a dataset of automatically aligned sentence pairs,
with a scaled threshold greater than 0.45.1  In addition, around 51.5 million potential
matches, with a scaled score below 0.45, are pruned
from the dataset.

\begin{table*}[tb]
\centering
\caption{\label{good} Annotated examples: the matching regions for partial and good partial are italicized (table from \cite{sim})}
\begin{tabular}{c|c c}\hline
 & standard & simple \\ \hline
good & \pbox{20cm}{Apple sauce or applesauce is a puree \\ made of apples.} & \pbox{20cm}{Applesauce (or applesauce) is a sauce that is \\made from stewed or mashed apples.}\\ \hline
\pbox{20cm}{good \\partial} & \pbox{20cm}{Commercial versions of applesauce \\ are really available in supermarkets.} & \pbox{20cm}{It is easy to make at home, and \textit{it is also sold} \\ \textit{already made in supermarkets as a common food}}.\\ \hline
partial & \pbox{20cm}{\textit{Applesauce} is a sauce that is \textit{made from}\\ stewed and mashed apples.} & \pbox{20cm}{\textit{Applesauce is made} by cooking down apples \\ with water or apple cider to the desired level.}\\ \hline
\end{tabular}
\end{table*}

In particular, the dataset use their best case method (structured Wordnet) to align sentences from 22k standard and simple articles, which were downloaded in April 2014. 
Based on the precision-recall tested on annotated data, a scaled threshold
of 0.67 (P = 0.798, R = 0.599, F1 = 0.685)
for good matches is selected, which gives the highest f1 score. The selected
thresholds yield around 150k good matches,
130k good partial matches, and 110k uncategorized
matches.

In order to provide more discriminitave information to the model, we trace back to the article that contains the good sentence pairs and randomly select 9 sentences that have similarity score lower than 0.67 (the threshold that generate the wikipedia dump) as negative samples.
Each sample contains one simple sentence, one corresponding matched standard sentence (positive), and 9 randomly selected standard sentences (negative). 

\subsection{Model variations}

\begin{description}
\item[CLSM9-letter] \hfill \\ Use letter trigram representation for each word. For the letter trigram that have frequency less than 5, replace the token as $<UNK>$. The total letter trigram size is 11024.This model is trained by randomly select 9 negative samples from the same article as positive sample, as described above.
\item[CLSM12-letter] \hfill \\ Use letter trigram representation for each word. This model is trained by randomly select 9 negative samples from the same article as positive sample and 3 negative samples randomly selected from different articles.
\item[CLSM9-word2vec] \hfill \\ Use Skip-gram \cite{word2vec} to represent each word. The word embeddings are fixed while training. If the word frequecy is less than 5, the word token is replaced by $<UNK>$. The total vocabulary size is 108874.  100, 150, 200, 250 dimension of word embeddings are tested. Word embeddings are pre-trained by all Wikipedia articles, including bothe simple and standard articles.
\item[CLSM9-word2vec-ft] \hfill \\ Use Skip-gram  to represent each word. The word embeddings are fine tuned during training. 100, 150, 200, 250 dimension of word embeddings are tested.
 \end{description}

In order to maintain consistency, the context window is kept 3 (word-trigram) across different models.
Hidden layer dimension use exactly the same parameters as \cite{clsm}. The dimension of convolutional layer and semantic layer are 300 and 120 respectively.
 Learning rate are tuned from 0.001 to 0.1 for different models. 
The model is trained by stochastic gradient descent.

\subsection{Experimental Results} 
The performance of the four different models tested on automaticly aligned Wikipedia sentence pairs are shown in Table \ref{result1}.
The whole data set is split into three parts: 150k samples for training, 5k samples for validation and 5k for testing. The parameters with the best result are used for aligning annotated dataset. The validation and testing task is the identify the best matched pair given the target simple/standard sentence and the 9 negative samples (select 1 out of 10 sentences). For CLSM12-letter model, the task is to select 1 out of 13 sentences, the within article negative sentences remains the same across different experiments.

\begin{table*}[htbp]
\centering
\caption{\label{result1} Experimental results for automatic aligned Wikipedia sentence pairs}
\begin{tabular}{c|c|c}\hline
Model & validation accuracy (\%) & test accuracy (\%)\\ \hline
CLSM9-letter & $\bm{93.8}$ & $\bm{93.1}$ \\ \hline
CLSM12-letter & 93.2 & 92.8 \\ \hline
CLSM9-word2vec (100 dimension) & 92.3 & 92.8\\ \hline
CLSM9-word2vec (150 dimension) & 92.3 & 92.1\\ \hline
CLSM9-word2vec (200 dimension) & 92.7 & 92.4\\ \hline
CLSM9-word2vec (250 dimension) & 92.1 & 92.7\\ \hline
CLSM9-word2vec-ft (100 dimension) & 92.3 & 92.5\\ \hline
CLSM9-word2vec-ft (150 dimension) & 92.6 & 92.7\\ \hline
CLSM9-word2vec-ft (200 dimension) & 92.1 & 92.7\\ \hline
CLSM9-word2vec-ft (250 dimension) & 92.6 & 92.2\\ \hline
\end{tabular}
\end{table*}

From Table \ref{result1}, we could see letter trigram based models (CLSM9-letter and CLSM12-letter) give the best result. Word2vec initialiation and fine tuning are around the same performance. One possible reason is the data for pre-training word embedding is similar to the data training CLSM (CLSM data is a subset). Even though letter-trigram based models have much more parameters  ($30k \times 300$) than word embedding based models on convolution layer ($30k \times 300$  for 100 dimension word embedding), the performance is not significantly improved. The computational time for CLSM9-letter on all Wikipedia pairs are over 72H, while the training time for word2vec embedding initialization is less than 4H.   

In order to test the performance of sentence level alignment, we use best parameters in Table \ref{result1} to test the annotated dataset. The similarity score is calculated using Eq. \ref{similarity}. Then we do greedy sequence-level alignment based on the extracted score as described in Section \ref{greedy}. The precision, recall and f1 score are in Table \ref{result2}. The baseline methods are described in Section \ref{baseline}. Note that Structured WordNet is the method that generates our training data for CLSM. 

\begin{table*}[htbp]
\centering
\caption{\label{result2} Experimental results for aligning annotated data}
\begin{tabular}{c|c|c|c}\hline
Model & precision & recall & f1 \\ \hline
WordNet (Baseline) & 0.80 & 0.60 & 0.69 \\ \hline
CLSM9-letter & 0.80 & 0.57 & 0.67 \\ \hline
CLSM12-letter & 0.81 & 0.55 & 0.66 \\ \hline
CLSM9-word2vec (100 dimension) & 0.73 & 0.56 & 0.63\\ \hline
CLSM9-word2vec-ft (150 dimension) & 0.76 & 0.56 & 0.64\\ \hline

\end{tabular}
\end{table*}

From Table \ref{result2}, CLSM give us a close performance as baseline method with the same precision and sligtly lower recall. This indicates that our model could learn the underlying relations between sentence pairs in training data. 
We then tried rescoring similarity scores generated from CLSM9-letter model and WordNet, we further get a gain with 0.71 f1 score. 
Since the training data is noisy, we could get a higher performance, therefore our next step is to use part of annotated data to adapt models trained from automatically generated data.

\section{Analysis}
\begin{figure}[tb]
\centering
\includegraphics[width=8cm]{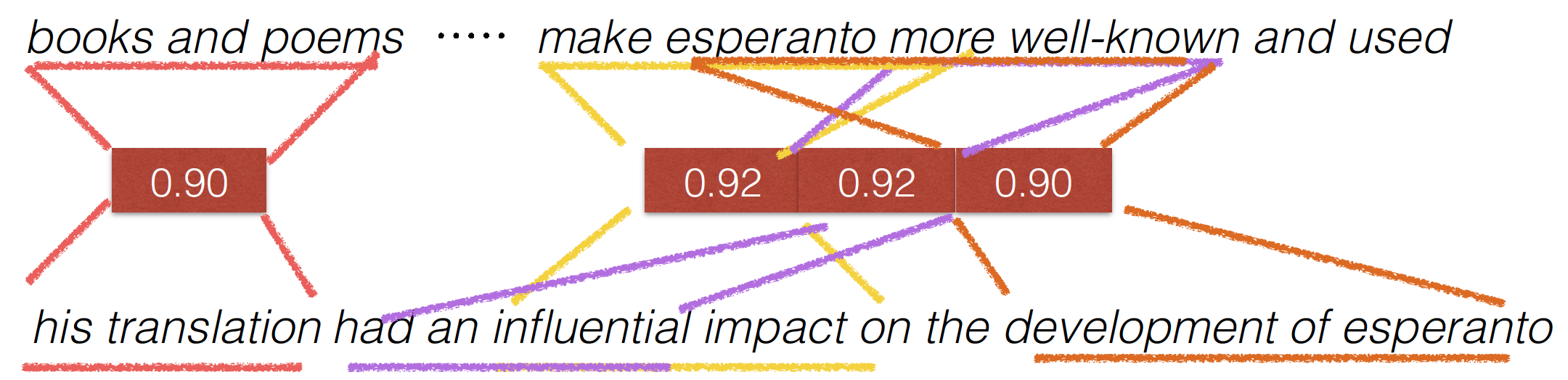}
\caption{{\it An example of word-level semantic matching in two matched sentence pairs}}
\label{2}
\end{figure}

According to \cite{clsm}, different words with related semantic meanings activate the similar set of neurons, resulting to a high overall matching score. This fact could give us inspiration to further pinpointing phrase-level matches within sentence pairs. Therefore, for each matched sentence pairs, we first project two sentences to max-pooling layer. Then, we evaluate the activation values of neurons at the max-pooling layer, and show the indice of the neurons that have high activation values of both sentences. After that, we trace back to the words that win these neurons in both sentences. An example of high activation neurons and the words in sentence pairs are shown in Figure \ref{2}. The simple sentence is `The books and poems that he changed into esperanto from other languages helped to make esperanto more well-known and used'. The corresponding standard sentence is `His translation had a influential impact on the development of esperanto into a language of literature'. From Figure \ref{2}, `books and poems' matches with `his translation', while `make esperanto more well-known and used' matches with `had an influential impact on the development of esperanto'. In this example, though there is no overlap between the two phrases, they both have high activation values at a similar set of neurons, thus lead to a sentence-level match in the semantic space. In future work, we could do phrase-level alignment based on this property of CLSM.

\section{Conclusion and future work}
In this project, we implemented a deep learning architecture called CLSM, the higher layer of the entire architecture is effective in generating latent semantic vector representations. CLSM is used for sentence level alignment in simple and standard Wikipedia. 
CLSM gives good performance in matching standard and simple sentences. However, since the training data we use is automatically generated and is thus noisy. Our next step is to adapt the model to annotated data.

\bibliographystyle{IEEEbib}
\small
\bibliography{references}

\begin{thebibliography}{}

\bibitem[\protect\citename{Aho and Ullman}1972]{Aho:72}
Alfred~V. Aho and Jeffrey~D. Ullman.
\newblock 1972.
\newblock {\em The Theory of Parsing, Translation and Compiling}, volume~1.
\newblock Prentice-Hall, Englewood Cliffs, NJ.

\bibitem[\protect\citename{{American Psychological Association}}1983]{APA:83}
{American Psychological Association}.
\newblock 1983.
\newblock {\em Publications Manual}.
\newblock American Psychological Association, Washington, DC.

\bibitem[\protect\citename{Chandra \bgroup et al.\egroup }1981]{Chandra:81}
Ashok~K. Chandra, Dexter~C. Kozen, and Larry~J. Stockmeyer.
\newblock 1981.
\newblock Alternation.
\newblock {\em Journal of the Association for Computing Machinery},
  28(1):114--133.

\bibitem[\protect\citename{Gusfield}1997]{Gusfield:97}
Dan Gusfield.
\newblock 1997.
\newblock {\em Algorithms on Strings, Trees and Sequences}.
\newblock Cambridge University Press, Cambridge, UK.

\end{thebibliography}


\begin{thebibliography}{10}

\bibitem{luan2017scienceie}
Yi~Luan, Mari Ostendorf, and Hannaneh Hajishirzi,
\newblock ``Scientific information extraction with semi-supervised neural
  tagging,''
\newblock in {\em Proc.\ Conf. Empirical Methods Natural Language Process.
  (EMNLP)}, 2017.

\bibitem{callison}
Chris Callison-Burch and Miles Osborne,
\newblock ``Bootstrapping parallel corpora,''
\newblock in {\em Proceedings of the HLT-NAACL 2003 Workshop on Building and
  using parallel texts: data driven machine translation and beyond-Volume 3}.
  Association for Computational Linguistics, 2003, pp. 44--49.

\bibitem{fung}
Pascale Fung and Percy Cheung,
\newblock ``Mining very-non-parallel corpora: Parallel sentence and lexicon
  extraction via bootstrapping and e,''
\newblock in {\em Proceedings of the 2004 Conference on Empirical Methods in
  Natural Language Processing}, 2004.

\bibitem{yatskar}
Mark Yatskar, Bo~Pang, Cristian Danescu-Niculescu-Mizil, and Lillian Lee,
\newblock ``For the sake of simplicity: Unsupervised extraction of lexical
  simplifications from wikipedia,''
\newblock in {\em Human Language Technologies: The 2010 Annual Conference of
  the North American Chapter of the Association for Computational Linguistics}.
  Association for Computational Linguistics, 2010, pp. 365--368.

\bibitem{zhu}
Zhemin Zhu, Delphine Bernhard, and Iryna Gurevych,
\newblock ``A monolingual tree-based translation model for sentence
  simplification,''
\newblock in {\em Proceedings of the 23rd international conference on
  computational linguistics}. Association for Computational Linguistics, 2010,
  pp. 1353--1361.

\bibitem{sim}
William Hwang, Hannaneh Hajishirzi, Mari Ostendorf, and Wei Wu,
\newblock ``Aligning sentences from standard wikipedia to simple wikipedia,''
\newblock in {\em Proceedings of the 2015 Conference of the North American
  Chapter of the Association for Computational Linguistics: Human Language
  Technologies}, 2015, pp. 211--217.

\bibitem{image}
Yann LeCun, Yoshua Bengio, et~al.,
\newblock ``Convolutional networks for images, speech, and time series,''
\newblock {\em The handbook of brain theory and neural networks}, vol. 3361,
  no. 10, pp. 1995, 1995.

\bibitem{speech}
Ossama Abdel-Hamid, Abdel-rahman Mohamed, Hui Jiang, and Gerald Penn,
\newblock ``Applying convolutional neural networks concepts to hybrid nn-hmm
  model for speech recognition,''
\newblock in {\em Acoustics, Speech and Signal Processing (ICASSP), 2012 IEEE
  International Conference on}. IEEE, 2012, pp. 4277--4280.

\bibitem{clsm}
Yelong Shen, Xiaodong He, Jianfeng Gao, Li~Deng, and Gr{\'e}goire Mesnil,
\newblock ``A latent semantic model with convolutional-pooling structure for
  information retrieval,''
\newblock in {\em Proceedings of the 23rd ACM International Conference on
  Conference on Information and Knowledge Management}. ACM, 2014, pp. 101--110.

\bibitem{match}
Baotian Hu, Zhengdong Lu, Hang Li, and Qingcai Chen,
\newblock ``Convolutional neural network architectures for matching natural
  language sentences,''
\newblock in {\em Advances in neural information processing systems}, 2014, pp.
  2042--2050.

\bibitem{luan2016multiplicative}
Yi~Luan, Yangfeng Ji, Hannaneh Hajishirzi, and Boyang Li,
\newblock ``Multiplicative representations for unsupervised semantic role
  induction,''
\newblock in {\em Proc.\ Annu. Meeting Assoc. for Computational Linguistics
  (ACL)}, 2016.

\bibitem{luan2016lstm}
Yi~Luan, Yangfeng Ji, and Mari Ostendorf,
\newblock ``{LSTM} based conversation models,''
\newblock {\em arXiv preprint arXiv:1603.09457}, 2016.

\bibitem{word2veccnn}
Yoon Kim,
\newblock ``Convolutional neural networks for sentence classification,''
\newblock {\em arXiv preprint arXiv:1408.5882}, 2014.

\bibitem{word2vec}
Tomas Mikolov, Kai Chen, Greg Corrado, and Jeffrey Dean,
\newblock ``Efficient estimation of word representations in vector space,''
\newblock {\em arXiv preprint arXiv:1301.3781}, 2013.

\bibitem{luan2018multi}
Yi~Luan, Luheng He, Mari Ostendorf, and Hannaneh Hajishirzi,
\newblock ``Multi-task identification of entities, relations, and coreference
  for scientific knowledge graph construction,''
\newblock {\em arXiv preprint arXiv:1808.09602}, 2018.

\bibitem{luan2015efficient}
Yi~Luan, Shinji Watanabe, and Bret Harsham,
\newblock ``Efficient learning for spoken language understanding tasks with
  word embedding based pre-training.,''
\newblock in {\em INTERSPEECH}, 2015.

\bibitem{luan2014relating}
Yi~Luan, Richard Wright, Mari Ostendorf, and Gina-Anne Levow,
\newblock ``Relating automatic vowel space estimates to talker
  intelligibility,''
\newblock in {\em Proc.\ Conf. Int. Speech Communication Assoc. (INTERSPEECH)},
  2014.

\bibitem{luan2017multi}
Yi~Luan, Chris Brockett, Bill Dolan, Jianfeng Gao, and Michel Galley,
\newblock ``Multi-task learning for speaker-role adaptation in neural
  conversation models,''
\newblock {\em Proc. Int. Conf. Joint Conference on Natural Language Processing
  (IJCNLP)}, 2017.

\bibitem{DSSM}
Po-Sen Huang, Xiaodong He, Jianfeng Gao, Li~Deng, Alex Acero, and Larry Heck,
\newblock ``Learning deep structured semantic models for web search using
  clickthrough data,''
\newblock in {\em Proceedings of the 22nd ACM international conference on
  Conference on information \& knowledge management}. ACM, 2013, pp.
  2333--2338.

\bibitem{socher}
Richard Socher, Jeffrey Pennington, Eric~H Huang, Andrew~Y Ng, and
  Christopher~D Manning,
\newblock ``Semi-supervised recursive autoencoders for predicting sentiment
  distributions,''
\newblock in {\em Proceedings of the conference on empirical methods in natural
  language processing}. Association for Computational Linguistics, 2011, pp.
  151--161.

\bibitem{collobert}
Ronan Collobert, Jason Weston, L{\'e}on Bottou, Michael Karlen, Koray
  Kavukcuoglu, and Pavel Kuksa,
\newblock ``Natural language processing (almost) from scratch,''
\newblock {\em Journal of Machine Learning Research}, vol. 12, no. Aug, pp.
  2493--2537, 2011.

\bibitem{luan2018uwnlp}
Yi~Luan, Mari Ostendorf, and Hannaneh Hajishirzi,
\newblock ``The {UWNLP} system at {SemEval-2018 Task 7}: Neural relation
  extraction model with selectively incorporated concept embeddings,''
\newblock in {\em Proc.\ Int.\ Workshop on Semantic Evaluation (SemEval)},
  2018, pp. 788--792.

\bibitem{wordnet}
Michael Mohler and Rada Mihalcea,
\newblock ``Text-to-text semantic similarity for automatic short answer
  grading,''
\newblock in {\em Proceedings of the 12th Conference of the European Chapter of
  the Association for Computational Linguistics}. Association for Computational
  Linguistics, 2009, pp. 567--575.

\bibitem{parser}
Marie-Catherine De~Marneffe, Bill MacCartney, Christopher~D Manning, et~al.,
\newblock ``Generating typed dependency parses from phrase structure parses,''
\newblock in {\em Proceedings of LREC}. Genoa Italy, 2006, vol.~6, pp.
  449--454.

\end{thebibliography}

\end{document}